\documentclass[sigconf,screen,nonacm]{acmart}
\usepackage{fontspec}

\usepackage{multirow}
\usepackage{array}
\newcolumntype{P}[1]{>{\centering\arraybackslash}p{#1}}
\usepackage{booktabs}
\usepackage[flushleft]{threeparttable}
\usepackage{makecell}
\usepackage{graphicx}
\usepackage{siunitx}
\usepackage{enumitem}
\usepackage{threeparttable}
\usepackage{xcolor}
\usepackage{pifont}
\usepackage{makecell}
\usepackage{pifont}

\newcommand{\cmark}{\textbf{\ding{51}}}
\newcommand{\xmark}{\textcolor{gray}{\ding{55}}}

\AtBeginDocument{%
  }
\usepackage[table]{xcolor} 
\usepackage{color}
\definecolor{rowcolor}{rgb}{0.898, 0.949, 0.969}
\renewcommand\footnotetextcopyrightpermission[1]{}
\setcopyright{acmlicensed}
\copyrightyear{2018}
\acmYear{2018}
\acmDOI{XXXXXXX.XXXXXXX}
\acmConference[Conference acronym 'XX]{Make sure to enter the correct
  conference title from your rights confirmation email}{June 03--05,
  2018}{Woodstock, NY}
\acmISBN{978-1-4503-XXXX-X/2018/06}

\acmSubmissionID{123-A56-BU3}

\begin{document}

\title{PaveBench: A Versatile Benchmark for Pavement Distress Perception and Interactive Vision-Language Analysis}

\author{Dexiang Li}
\email{lambert1862@163.com}
\orcid{0009-0008-9663-1953}
\affiliation{%
  \institution{Harbin Institute of Technology}
  \city{Shenzhen}
  \country{China}
}

\author{Zhenning Che}
\email{chezhenning1@gmail.com}
\orcid{0009-0004-8520-7701}
\affiliation{%
  \institution{Harbin Institute of Technology}
  \city{Shenzhen}
  \country{China}
}

\author{Haijun Zhang}
\email{hjzhang@hit.edu.cn}
\orcid{0000-0002-1648-0227}
\authornote{Corresponding authors.}
\affiliation{%
  \institution{Harbin Institute of Technology}
  \city{Shenzhen}
  \country{China}
}

\author{Dongliang Zhou}
\email{zhou-dongliang@outlook.com}
\orcid{0000-0003-0361-8597}
\authornotemark[1]
\affiliation{%
  \institution{Tianjin University}
  \city{Tianjin}
  \country{China}
}

\author{Zhao Zhang}
\email{cszzhang@gmail.com}
\orcid{0000-0002-5703-7969}
\affiliation{%
  \institution{Hefei University of Technology }
  \city{Hefei}
  \country{China}
}

\author{Yahong Han}
\email{yahong@tju.edu.cn}
\orcid{0000-0003-2768-1398}
\affiliation{%
  \institution{Tianjin University}
  \city{Tianjin}
  \country{China}
}


\begin{abstract}
Pavement condition assessment is essential for road safety and maintenance. Existing research has made significant progress. However, most studies focus on conventional computer vision tasks such as classification, detection, and segmentation. In real-world applications, pavement inspection requires more than visual recognition. It also requires quantitative analysis, explanation, and interactive decision support. Current datasets are limited. They focus on unimodal perception. They lack support for multi-turn interaction and fact-grounded reasoning. They also do not connect perception with vision-language analysis. To address these limitations, we introduce \textit{PaveBench}, a large-scale benchmark for pavement distress perception and interactive vision-language analysis on real-world highway inspection images. PaveBench supports four core tasks: classification, object detection, semantic segmentation, and vision-language question answering. It provides unified task definitions and evaluation protocols. On the visual side, PaveBench provides large-scale annotations and includes a curated hard-distractor subset for robustness evaluation. It contains a large collection of real-world pavement images. On the multimodal side, we introduce PaveVQA, a real-image question answering (QA) dataset that supports single-turn, multi-turn, and expert-corrected interactions. It covers recognition, localization, quantitative estimation, and maintenance reasoning. We evaluate several state-of-the-art methods and provide a detailed analysis. We also present a simple and effective agent-augmented visual question answering framework that integrates domain-specific models as tools alongside vision-language models. The dataset is available at: \url{https://huggingface.co/datasets/MML-Group/PaveBench}.
\end{abstract}





\maketitle

\section{Introduction}

\begin{figure}[t] 
  \centering
  \includegraphics[width=1\linewidth]{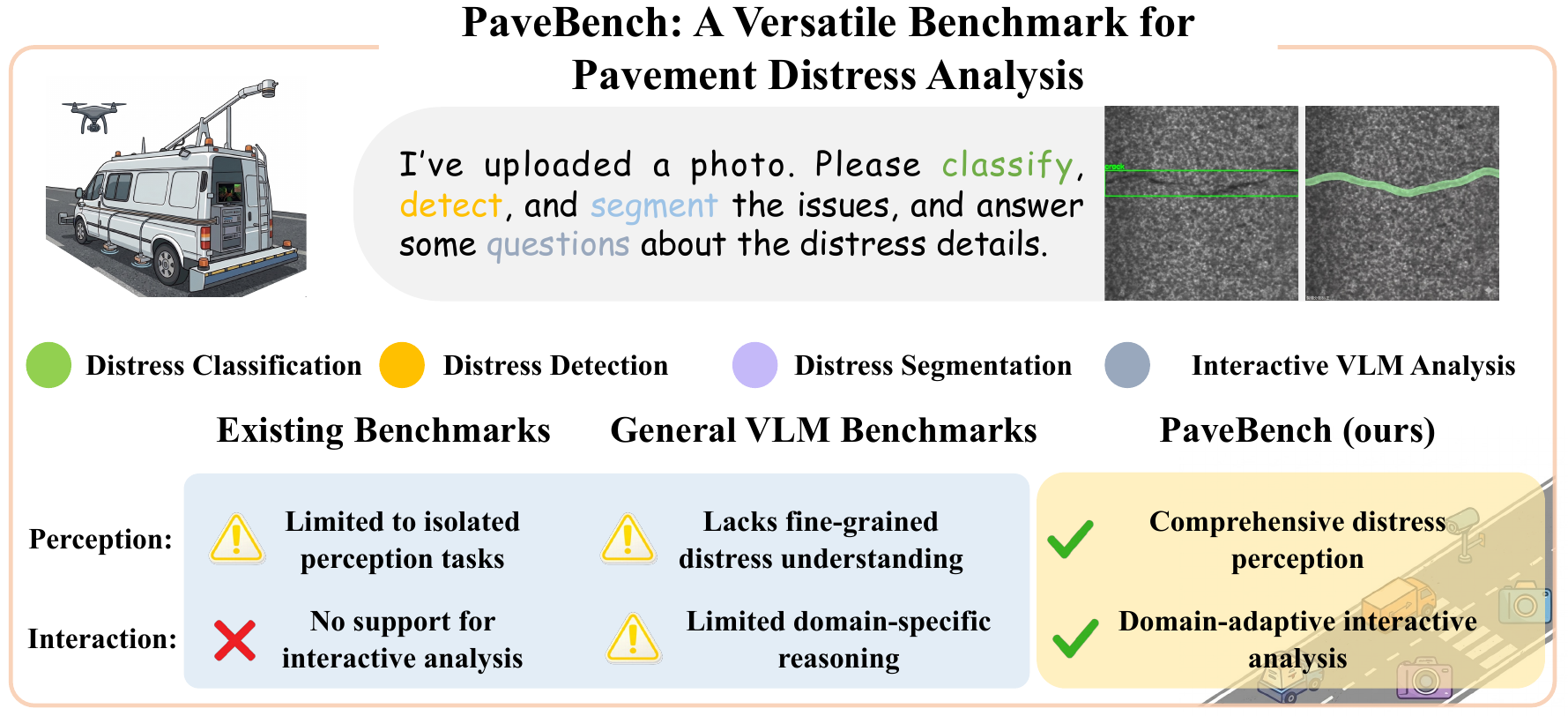} 
  \caption{Comparison between existing benchmarks, general VLM benchmarks, and PaveBench. PaveBench unifies perception and interactive vision-language analysis, providing comprehensive and domain-aware pavement distress understanding.}
  \label{fig:overview4cmp}
\end{figure}

Road networks are essential to modern society. Their pavement condition directly affects safety, efficiency, and maintenance costs. As road infrastructure ages and traffic demand continues to grow, scalable and reliable pavement inspection is becoming increasingly important. Existing research has made considerable progress, but it remains centered mainly on conventional computer vision tasks such as classification, detection, and segmentation. In real-world settings, however, pavement inspection requires not only visual recognition but also quantitative analysis, explanation, and interactive decision support. As a result, pavement assessment shall extend beyond a purely visual recognition problem.

Datasets are fundamental to pavement distress perception research, as they define task scope and guide research direction. However, most existing datasets and studies focus on visual recognition. In particular, they do not support richer analytical or interactive capabilities. Early datasets, such as CFD~\cite{shi2016crackforest} and CRACK500~\cite{yang2019feature}, are limited to single-type crack detection and segmentation. Later datasets, such as RDD2022~\cite{arya2024rdd2022}
, expand category coverage but do not support segmentation-level annotation. To unify multiple visual tasks, PaveDistress~\cite{liu2024pavedistress} was introduced, but it remains limited to a unimodal setting and does not support vision–language interaction.
At the same time, general vision-language models (VLMs) have made rapid progress in recent years, relying heavily on large-scale datasets such as COCO~\cite{lin2014microsoft} and LLaVA-Instruct~\cite{liu2023visual}. However, because these datasets are predominantly collected from the Internet, they provide limited coverage of the specialized concepts and fine-grained understanding required for pavement distress analysis. In response, RoadBench~\cite{xiao2026roadbench} serves as an early multimodal benchmark for this domain. Still, it relies on synthetic images, lacks segmentation-level annotations, and supports only coarse descriptions rather than rich question answering (QA). Overall, current datasets have three main limitations: (i) they do not support natural vision-language interaction on real-world acquired pavement images; (ii) they lack annotations for multi-turn dialogue and fact-grounded reasoning; and (iii) they do not provide a coherent data foundation for connecting specialized perception tasks with subsequent language-based analysis.

To address these limitations, we introduce \textit{PaveBench}, a versatile benchmark for pavement distress perception and interactive vision-language analysis on real-world highway inspection images. To our knowledge, it is the first benchmark in this domain to unify visual perception and multimodal reasoning, addressing the gap between traditional pavement distress methods and general VLMs shown in Fig.~\ref{fig:overview4cmp}. In particular, PaveBench covers four core tasks: classification, object detection, semantic segmentation, and vision-language question answering. It also provides unified task definitions and evaluation protocols to support consistent evaluation and facilitate future research. On the visual side, we provide large-scale annotations on top-down orthographic images, preserving geometric fidelity, retaining challenging hard-distractor cases. To the best of our knowledge, PaveBench covers the largest collection of real-world pavement images in this domain. On the multimodal side, we introduce PaveVQA, the largest real-image QA dataset for pavement distress analysis, supporting single-turn, multi-turn, and expert-refined dialogues across recognition, localization, quantitative estimation, and maintenance-oriented reasoning. Together, they establish a unified foundation for visually grounded multi-step diagnostic analysis.
For all four tasks, we conduct experiments on several state-of-the-art baselines and provide corresponding evaluation metrics and analyses. To further support the newly introduced analysis task, we present a simple and effective agent-augmented visual question answering (VQA) framework that natively integrates domain-specific visual tools with VLMs.

The main contributions of this paper can be summarized as follows:
\begin{itemize}
    \item We introduce PaveBench, a large-scale real-world benchmark for pavement distress analysis. It provides unified annotations for classification, detection, and segmentation, along with a curated hard-distractor subset for robustness evaluation in complex scenes.
    \item We build the largest real-image VQA benchmark for pavement distress inspection. It includes single-turn, multi-turn, and expert-corrected interactions and supports queries about the presence, type, location, severity, quantitative measurement, and maintenance recommendations for distress.
    \item We present a simple and effective agent-augmented VQA framework. It integrates domain-specific visual tools with VLMs, reduces numerical hallucinations, and improves the transparency of interactive diagnosis.
\end{itemize}

\section{Related Work}
In this section, we review related studies on pavement distress recognition datasets and vision-language datasets for interactive analysis, covering complementary aspects of perception and reasoning.

\textbf{Pavement Distress Recognition Datasets.} Pavement distress perception has long been driven by task-specific visual datasets. Early benchmarks, such as AigleRN~\cite{amhaz2015automatic}, CFD~\cite{shi2016crackforest}, CrackTree260~\cite{zou2012cracktree}, and CRACK500~\cite{yang2019feature}, mainly focus on crack segmentation and serve as benchmarks for fine-grained structural extraction. However, these datasets focus on single-category distress and thus only partially reflect real inspection scenarios. Later datasets extend to multi-category distress recognition and detection to better align with practical needs. In particular, GAPs~\cite{eisenbach2017get} introduces structured annotations for asphalt pavements, while the RDD series~\cite{maeda2019road,maeda2021rdd2020,arya2024rdd2022} expands both category coverage and dataset scale.
Nevertheless, these datasets are mostly based on oblique or street-view images and provide image-level or bounding-box annotations, which limit precise geometric analysis and pixel-level reasoning. More recent datasets, such as PaveDistress~\cite{liu2024pavedistress}, move toward high-resolution semantic segmentation. However, they remain confined to unimodal visual perception. In contrast, PaveBench couples pixel-level annotations with vision-language analysis, supporting both precise quantification and high-level diagnostic reasoning.

\textbf{Vision-Language Datasets for Interactive Analysis.} Recent progress in vision-language learning has been driven by large-scale instruction and QA datasets, including general-purpose resources such as LLaVA-Instruct~\cite{liu2023visual} and domain-specific benchmarks such as ChartQA~\cite{masry2022chartqa}, SLAKE~\cite{liu2021slake}, PMC-VQA~\cite{zhang2023pmcvqa}, and LLaVA-Med~\cite{li2023llavamed}. These works highlight the importance of domain-specific supervision for professional analysis. Building on this foundation, multimodal systems are evolving from passive answering to interactive reasoning and tool use, as demonstrated by frameworks such as ReAct~\cite{yao2022react}, Visual ChatGPT~\cite{wu2023visualchatgpt}, VisProg~\cite{gupta2023visual}, and ToolVQA~\cite{yin2025toolvqa}. However, despite these advances, the pavement domain still lacks a real-image benchmark that jointly supports dense perception, precise quantitative QA, and multi-turn interaction. While RoadBench~\cite{xiao2026roadbench} serves as an initial attempt, it relies on synthetic images and remains limited in both annotation granularity and interaction depth. In contrast, PaveBench unifies real-image perception annotations with single- and multi-turn, quantitative, and expert-corrected VQA, establishing a benchmark for visually grounded, interactive pavement analysis.

\section{PaveBench Dataset}

\begin{figure*}[h] 
  \centering
  \includegraphics[width=1\linewidth]{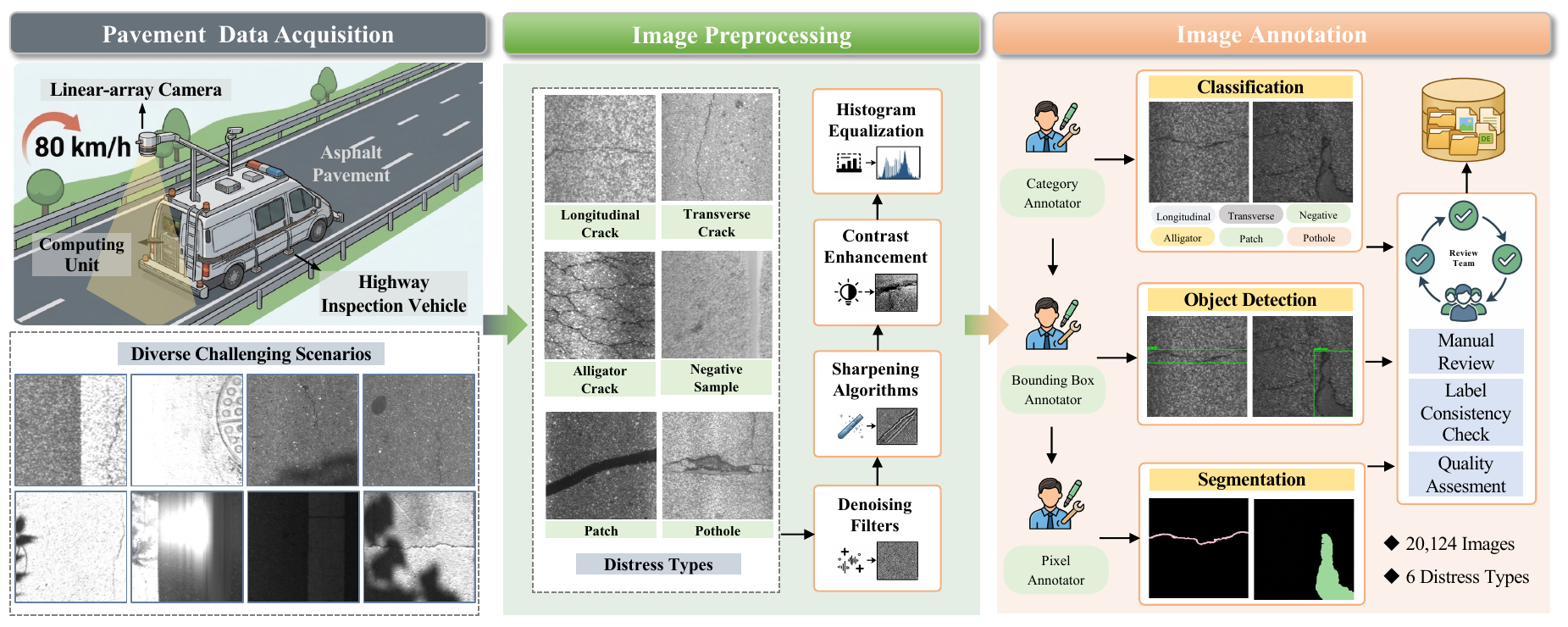}
  \caption{Data acquisition, annotation, and construction pipeline of the dataset for pavement distress perception tasks.}
  \label{fig2}
\end{figure*}

In this section, we present the construction of PaveBench. We describe the data acquisition process, the multi-task visual annotation pipeline, the identification of hard distractors, the construction of PaveVQA, and the overall dataset statistics and comparison with existing benchmarks.

\subsection{Data Acquisition and Collection} 
Raw pavement images were collected in Liaoning Province, China, using a highway inspection vehicle operating at 80 km/h. As shown in Fig.~\ref{fig2}(a), the system is equipped with a high-resolution line-scan camera that captures vertical orthographic (top-down) views during driving. This imaging setup preserves the geometric properties of pavement distress, such as crack width and length, and supports reliable downstream quantification. The collected data cover diverse and challenging scenarios, including shadows, stains, and varying lighting conditions. To improve visual quality, the raw continuous scans were further processed with a standard pipeline, including denoising, sharpening, contrast enhancement, and histogram equalization, as illustrated in Fig.~\ref{fig2}(b). These steps enhance the visibility of distress while reducing background noise, resulting in high-quality image patches for subsequent annotation and analysis.

\subsection{Multi-Task Annotation and Hard Distractor Curation}

This subsection introduces two key aspects of dataset construction: the multi-task annotation pipeline and the curation of hard distractors observed in real pavement scenes.

\textbf{Multi-Task Annotation.} As shown in Fig.~\ref{fig2}(c), all images were annotated through a hierarchical pipeline covering classification, detection, and segmentation. Classification labels were cross-validated by multiple annotators, and detection boxes were annotated using LabelMe\footnote{\url{https://labelme.io/}}. For pixel-level segmentation, four domain experts manually traced high-fidelity masks in Photoshop, averaging about ten minutes per image and up to one hour for complex alligator cracks. All annotations were further reviewed through a multi-stage expert verification process to ensure label consistency and accurate boundaries.

\textbf{Hard Distractor Curation.} During annotation, we found that real distresses often co-occur with visually confusing background patterns, such as pavement stains and shadows. These patterns can be easily mistaken for true distress regions and thus act as hard distractors. Instead of removing such cases, PaveBench explicitly categorizes and retains them as challenging samples for robustness evaluation. This design makes the benchmark more realistic and encourages models to distinguish structural distress from superficial visual noise.

\begin{figure*}[t] 
  \centering
  \includegraphics[width=1\linewidth]{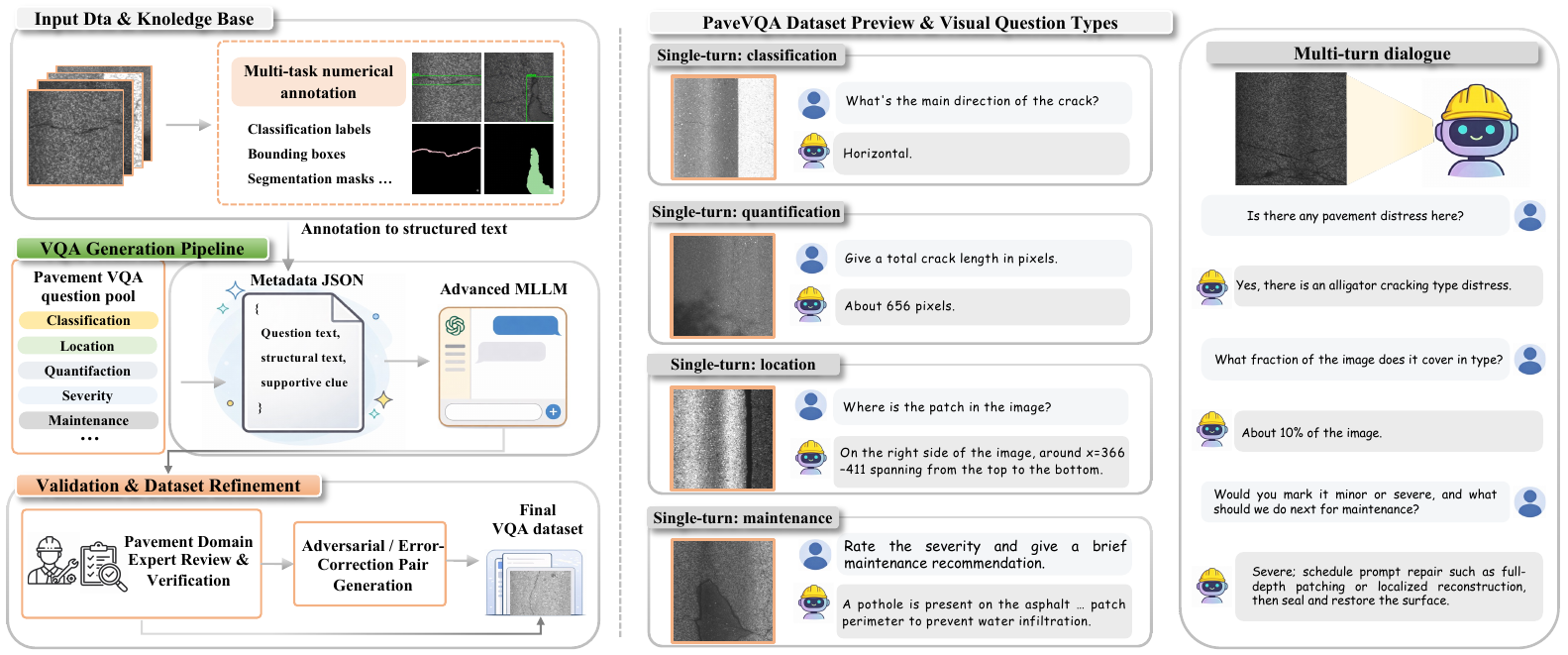} 
  \caption{Overview of PaveVQA, including its construction pipeline, visual question types, and multi-turn dialogue examples.}
  \label{fig3}
\end{figure*}

\subsection{PaveVQA Construction and Quality Control}

To construct PaveVQA at scale while maintaining data quality, we develop a structured pipeline for generating grounded and low-hallucination dialogues, as illustrated in Fig.~\ref{fig3}. The pipeline connects visual annotations, structured metadata, and large language model (LLM)-based generation into a unified framework.

\textbf{Question Design and Structured Metadata.} We first design a question pool that reflects practical inspection needs, including presence verification, classification, localization, quantitative analysis, severity assessment, and maintenance recommendation. To support reliable quantification, high-fidelity visual annotations are converted into structured JSON metadata. This metadata explicitly encodes geometric attributes, such as bounding box coordinates, pixel area, and skeleton length, providing verifiable evidence for downstream reasoning rather than relying on implicit visual representations.

\textbf{Dialogue Generation.} In the generation stage, raw images, structured metadata, and manually designed prompt templates are jointly fed into ChatGPT-5.2\footnote{\url{https://chatgpt.com/}} to produce diverse interactions. As shown in Fig.~\ref{fig3}, the resulting data cover multiple question types, including classification, quantification, localization, and maintenance-oriented reasoning. For each image, the pipeline generates ten single-turn questions and two rounds of multi-turn dialogue, yielding approximately 20 question-answer pairs.

\textbf{Quality Control.} To further improve reliability, we introduce negative queries about non-existent distresses, particularly for negative samples, encouraging the model to explicitly reject false premises. In addition, adversarial and error-correction pairs are constructed to expose potential reasoning failures. Finally, all generated samples are reviewed by pavement domain experts in a human-in-the-loop process to correct logical inconsistencies and ensure domain fidelity.

\begin{figure}[t] 
  \centering
  \includegraphics[width=1\linewidth]{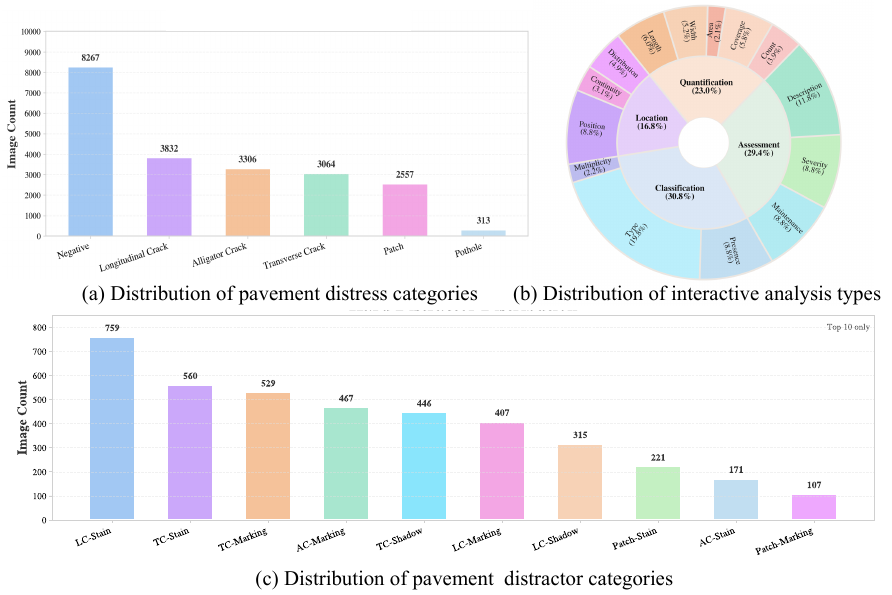} 
  \caption{Overview of data distributions in PaveBench. (a) Pavement distress category distribution. (b) Interactive analysis type distribution. (c) Fine-grained distress-condition distribution, showing a long-tailed and diverse dataset.}
  \label{fig4}
\end{figure}

 \subsection{Dataset Statistics and Analysis}

The statistics of PaveBench are summarized in Fig.~\ref{fig4}. As shown in Fig.~\ref{fig4}(a), the visual subset contains 20,124 high-resolution ($512 \times 512$) images. The class distribution is naturally imbalanced, reflecting real-world highway conditions where severe distresses are rare but critical. This imbalance poses a real challenge for models in detecting sparse yet important defects.
For the vision-language component, PaveVQA includes 32,160 question-answer pairs: 10,050 single-turn queries, 20,100 multi-turn interactions, and 2,010 error-correction pairs. To systematically evaluate different capabilities, the questions are organized into four primary tasks and 14 fine-grained sub-categories, as illustrated in Fig.~\ref{fig4}(b). This structured design enables evaluation across multiple levels, from basic perception to quantitative analysis and decision-oriented reasoning.
Moreover, Fig.~\ref{fig4}(c) shows the distribution of the top 10 pavement distractor categories. Here, AC, LC, and TC denote alligator crack, longitudinal crack, and transverse crack, respectively. The distractors often co-occur with these pavement distress in the same images and closely resemble real distress patterns, forming a challenging testbed for fine-grained discrimination and robustness evaluation. Overall, these statistics show that PaveBench reflects realistic pavement scenes and poses nontrivial challenges for both visual perception and multimodal reasoning.

\begin{table*}[t]
\centering
\caption{Comparison of existing pavement and crack benchmarks with PaveBench. Here, `language' denotes paired language supervision or image--text annotations.}
\label{tab:dataset_comparison_main}
\setlength{\tabcolsep}{7pt}
\renewcommand{\arraystretch}{1.15}
\resizebox{\textwidth}{!}{
\begin{tabular}{l c c c ccc cccc}
\toprule
\multirow{2}{*}{Dataset}
& \multirow{2}{*}{View}
& \multirow{2}{*}{Category}
& \multirow{2}{*}{\makecell{\#Images}}
& \multicolumn{3}{c}{Visual Perception}
& \multicolumn{4}{c}{Vision--Language Analysis} \\
\cmidrule(lr){5-7} \cmidrule(lr){8-11}
& & &
& Classification & Detection & Segmentation
& Single-turn & Multi-turn & \makecell{Quant. QA} & \makecell{Expert Corr.} \\
\midrule

AigleRN~\cite{amhaz2015automatic}
& Oblique
& Crack only
& 38
& \xmark & \xmark & \cmark
& \xmark & \xmark & \xmark & \xmark \\

CFD~\cite{shi2016crackforest}
& Oblique
& Crack only
& 118
& \xmark & \xmark & \cmark
& \xmark & \xmark & \xmark & \xmark \\

CRKWH100~\cite{liu2019deepcrack}
& Top-down
& Crack only
& 100
& \xmark & \xmark & \cmark
& \xmark & \xmark & \xmark & \xmark \\

CrackTree260~\cite{liu2019deepcrack}
& Oblique
& Crack only
& 260
& \xmark & \xmark & \cmark
& \xmark & \xmark & \xmark & \xmark \\

CrackLS315~\cite{zou2012cracktree}
& Top-down
& Crack only
& 315
& \xmark & \xmark & \cmark
& \xmark & \xmark & \xmark & \xmark \\

GAPs384~\cite{eisenbach2017get}
& Top-down
& Crack only
& 384
& \xmark & \xmark & \cmark
& \xmark & \xmark & \xmark & \xmark \\

CRACK500~\cite{yang2019feature}
& Oblique
& Crack only
& 500
& \xmark & \xmark & \cmark
& \xmark & \xmark & \xmark & \xmark \\

DeepCrack~\cite{liu2019deepcrack}
& Mixed
& Crack only
& 537
& \xmark & \xmark & \cmark
& \xmark & \xmark & \xmark & \xmark \\

Kaggle11k~\cite{kaggle11k}
& Mixed
& Crack only
& 11,298
& \xmark & \xmark & \cmark
& \xmark & \xmark & \xmark & \xmark \\

GAPs~\cite{eisenbach2017get}
& Top-down
& Multiple
& 1,969
& \xmark & \cmark & \xmark
& \xmark & \xmark & \xmark & \xmark \\

RDD2019~\cite{maeda2019road}
& Oblique
& Multiple
& 9,053
& \cmark & \cmark & \xmark
& \xmark & \xmark & \xmark & \xmark \\

RDD2020~\cite{maeda2021rdd2020}
& Oblique
& Multiple
& 26,620
& \cmark & \cmark & \xmark
& \xmark & \xmark & \xmark & \xmark \\


PID~\cite{majidifard2020pavement}
& Mixed
& Multiple
& 7,237
& \cmark & \cmark & \xmark
& \xmark & \xmark & \xmark & \xmark \\

RDD2022~\cite{arya2024rdd2022}
& Oblique
& Multiple
& 47,420
& \xmark & \cmark & \xmark
& \xmark & \xmark & \xmark & \xmark \\

PaveDistress~\cite{liu2024pavedistress}
& Top-down
& Multiple
& 6,032
& \cmark & \cmark & \cmark
& \xmark & \xmark & \xmark & \xmark \\

RoadBench~\cite{xiao2026roadbench}
& Oblique
& Multiple
& 100,000 \textsuperscript{\textdagger}
& \xmark & \cmark & \xmark
& \cmark & \xmark & \cmark & \xmark \\

\midrule
\rowcolor{gray!15}
PaveBench (\textbf{ours})
& Top-down
& Multiple
& 20,124
& \cmark & \cmark & \cmark
& \cmark & \cmark & \cmark & \cmark \\

\bottomrule
\end{tabular}
}
\begin{tablenotes}
            {
            \item \textsuperscript{\textdagger} \textit{Note:} All images in RoadBench are synthetically generated rather than collected from real-world scenes.}
\end{tablenotes}

\end{table*}

\subsection{Comparison with Existing Benchmarks}

Table~\ref{tab:dataset_comparison_main} compares PaveBench with representative pavement datasets. Early benchmarks such as CFD and CRACK500 focus on single-distress segmentation. Later datasets, including the RDD series, expand to multiple distress categories but primarily provide bounding box annotations from oblique views, which limits precise geometric analysis. More recent efforts attempt to extend task coverage. PaveDistress unifies classification, detection, and segmentation under a top-down setting, but remains limited to unimodal perception. RoadBench introduces language supervision, yet relies on synthetic imagery and does not support fine-grained segmentation or interactive dialogue.

In contrast, PaveBench is designed as a unified benchmark that connects visual perception with vision-language analysis. It provides 20,124 real top-down images with consistent annotations for classification, detection, and segmentation, enabling geometry-aware evaluation. Beyond visual tasks, it integrates multimodal capabilities, including quantitative question answering, multi-turn dialogue, and expert-verified error correction. This combination allows PaveBench to support both precise visual understanding and interactive reasoning within a single framework.

\begin{table}[t]
  \centering
    \caption{Comparison of classification performance on PaveBench.}
    \label{tab:classification}
    \resizebox{0.8\linewidth}{!}{
    \begin{tabular}{l cccc}
      \toprule
      \multirow{2}{*}{Method} & \multicolumn{4}{c}{Metrics (\%)} \\
      \cmidrule(lr){2-5}
      & Acc. ($\uparrow$) & Prec. ($\uparrow$) & Rec. ($\uparrow$) & F1 ($\uparrow$) \\
      \midrule
      ConvNeXt v2~\cite{woo2023convnext}   & \underline{91.19} & \underline{91.08} & \underline{91.19} & \underline{91.04} \\
      FASTERVIT~\cite{hatamizadeh2023fastervit}  & 87.36 & 87.40 & 87.36 & 87.03 \\
      TinyNeXt~\cite{zeng2025efficient}   & 90.78 & 90.74 & 90.88 & 90.71 \\
      LSNet~\cite{wang2025lsnet}      & 88.80 & 88.66 & 88.78 & 88.54 \\
      OverLoCK-T~\cite{lou2025overlock} & \textbf{93.81} & \textbf{93.81} & \textbf{93.81} & \textbf{93.76} \\
      \bottomrule
    \end{tabular}}
\end{table}

\section{Benchmark and Experiments}
In this section, we evaluate PaveBench on visual perception and multimodal VQA tasks. We first describe the evaluation protocol in Section \ref{sec:eva_pro}, then report results on classification, detection, and segmentation in Section \ref{sec:vis-perc}, followed by VQA evaluation in Section \ref{sec:mm-vqa}. Finally, we present an agent-augmented framework that integrates perception models with VLMs for more reliable analysis in Section \ref{sec:exp-agent}.

\subsection{Evaluation Protocol}
\label{sec:eva_pro}

We evaluate PaveBench under three complementary settings: visual perception, interactive VQA, and agent-augmented VQA. The visual benchmark covers image-level classification, instance-level detection, and pixel-level segmentation on real-world pavement images. The VQA benchmark assesses VLMs on distress-oriented queries. We further introduce an agent-augmented setting in which VLMs are equipped with specialized visual tools to produce strictly visually grounded responses.

\textbf{Visual Perception Metrics.} Following standard evaluation protocols~\cite{deng2009imagenet}, we assess image classification using top-1 accuracy (Acc.), macro-averaged precision (Prec.), recall (Rec.), and F1-score. Object detection is evaluated using standard COCO metrics~\cite{lin2014microsoft}, including mAP, AP$_{50}$, AP$_{75}$, and average recall (AR). For semantic segmentation, we measure morphological fidelity using mean precision (mPrec.), mean recall (mRec.), mean F1 (mF1), and mIoU~\cite{cordts2016cityscapes}.

\textbf{Multimodal VQA Metrics.} To evaluate reasoning and quantification capabilities, we design a dual-metric evaluation scheme for VQA tasks. For strict numerical and factual queries, we adopt task-specific metrics based on answer format: classification accuracy for categorical queries on distress presence and type; localization token-F1 for short-text spatial descriptions; and segmentation quantification MAE for pixel-level numerical estimation (e.g., crack length and area). For descriptive responses, including severity assessment and maintenance recommendation, we adopt standard text generation metrics, including ROUGE-L~\cite{lin2004rouge}, BLEU~\cite{papineni2002bleu}, METEOR~\cite{banerjee2005meteor}, and BERTScore~\cite{zhang2019bertscore}. The same evaluation protocol is applied to both fine-tuned VLMs and the agent-augmented setting.

\begin{table}[t]
  \centering
    \caption{Comparison of detection performance on PaveBench.}
    \label{tab:detection}
    \resizebox{0.8\linewidth}{!}{
    \begin{tabular}{l cccc}
      \toprule
      \multirow{2}{*}{Method} & \multicolumn{4}{c}{Metrics (\%)} \\
      \cmidrule(lr){2-5}
      & mAP ($\uparrow$) & AP$_{50}$ ($\uparrow$) & AP$_{75}$ ($\uparrow$) & AR ($\uparrow$) \\
      \midrule
      YOLO2026~\cite{sapkota2025yolo26} & 64.52 & \underline{83.60} & \underline{68.53} & 74.76 \\
          RemDet~\cite{li2025remdet}   & \underline{68.30} & 73.80 & 62.10 & 77.77 \\
      MI-DETR~\cite{nan2025mi}  & 64.80 & 72.96 & 66.96 & \underline{87.70} \\
      DEIM~\cite{huang2025deim}     & \textbf{71.84} & \textbf{85.87} & \textbf{76.91} & \textbf{89.82} \\
      \bottomrule
    \end{tabular}}
    \vspace{-0.5cm}
\end{table}

\subsection{Visual Perception Evaluation}
\label{sec:vis-perc}

We first evaluate PaveBench as a benchmark for visual perception to establish its value for understanding fundamental pavement distress. As summarized in Tables~\ref{tab:classification}, \ref{tab:detection}, and \ref{tab:segmentation}, modern architectures achieve competitive results across classification, detection, and segmentation, showing that PaveBench provides reliable supervision at the image, instance, and pixel levels. At the same time, the benchmark remains challenging. The best-performing models reach 92.27\% accuracy, 71.84\% mAP, and 76.0\% mIoU, leaving clear room for improvement in precise detection and fine-grained segmentation. This difficulty is largely due to complex real-world scenes, where pavement distress often co-occurs with hard distractors, such as shadows and stains. These factors introduce significant visual ambiguity, making it difficult to distinguish true structural distress from confusing background patterns. Overall, the results show that PaveBench serves as both a unified multi-task benchmark and a challenging testbed for robust pavement distress perception.

\begin{table}[t]
    \centering
    \caption{Comparison of segmentation performance on PaveBench.}
    \label{tab:segmentation}
    \resizebox{0.8\linewidth}{!}{
    \begin{tabular}{lcccc}
      \toprule
      \multirow{2}{*}{Method} & \multicolumn{4}{c}{Metrics (\%)} \\
      \cmidrule(lr){2-5}
      & mPrec. ($\uparrow$) & mRec. ($\uparrow$) & mF1 ($\uparrow$) & mIoU ($\uparrow$) \\
      \midrule
      DeepLabV3+~\cite{chen2018encoder}   & \underline{70.26} & 69.57 & 69.79 & 54.10 \\
      SegFormer~\cite{xie2021segformer}   & 69.69 & 66.61 & \underline{70.59} & \underline{55.44} \\
      SOSNet~\cite{liu2025sosnet}         & 70.02 & \underline{70.18} & 69.92 & 54.85 \\
      SCSegamba~\cite{liu2025scsegamba}   & \textbf{71.80} & \textbf{70.78} & \textbf{71.18} & \textbf{56.59} \\
      \bottomrule
    \end{tabular}
  }
\end{table}

\subsection{Multimodal VQA Evaluation}
\label{sec:mm-vqa}

We evaluate multimodal VQA performance on PaveBench using three vision-language models: Qwen2.5-VL~\cite{wang2023qwenvl}, DeepSeek-VL2~\cite{lu2024deepseekvl}, and LLaVA-OneVision~\cite{li2024llavaonevision}.
Although these models demonstrate strong general visual-language capabilities, they are not well aligned with the structured taxonomy, reasoning requirements, and quantitative demands of pavement distress analysis.
To address this gap, we apply low rank adaptation (LoRA)~\cite{hu2021lora} on the PaveVQA training set. This parameter-efficient tuning aligns model outputs with structured formats and improves their ability to produce consistent and task-relevant responses.
Table~\ref{tab:agent_metrics_comparison} reports the performance before and after fine-tuning. In the zero-shot setting, all models perform poorly on domain-specific queries, resulting in low classification accuracy and weak semantic scores. After LoRA adaptation, all models show consistent improvements across metrics. In particular, the reduction in quantification MAE indicates improved estimation of distress-related measurements. Gains in language metrics (e.g., ROUGE-L, BLEU, and METEOR) further suggest more coherent and task-aligned responses, enabling more reliable analysis and recommendation generation.

\subsection{Agent-Augmented VQA Framework}
\label{sec:exp-agent}
As a complementary approach to instruction fine-tuning, we introduce an agent-augmented framework to address the geometric and spatial limitations of general-purpose VLMs. Instead of encoding distress-related measurements in model parameters, the framework treats the VLM as an interactive controller that coordinates external tools. The VLM handles language understanding, while specialized models perform visual perception. This design reduces multimodal hallucinations.

In particular, our proposed framework integrates domain-specific models (Section \ref{sec:vis-perc}) into an interactive reasoning pipeline through tool-calling capabilities~\cite{yao2022react, wu2023visualchatgpt}. Given a user query, the VLM interprets the intent and decomposes it into executable steps. It then routes each subtask to a dedicated model, including OverLoCK-T for classification, DEIM for localization, and SCSegamba for segmentation. The visual outputs, such as bounding boxes and pixel masks, are converted into explicit geometric quantities. These quantities are then fed into the model context as textual evidence. 
The framework is interpretable because its intermediate results can be directly visualized and aligned with the generated responses. It also preserves dialogue history and visual states, which enables multi-turn, context-aware analysis from distress identification to quantitative assessment and decision support. Table~\ref{tab:agent_metrics_comparison} shows that the proposed agent-augmented framework achieves competitive performance on both numerical and language metrics. Importantly, this is achieved without additional parameter updates to the underlying VLMs. These results show that explicit tool-assisted reasoning can provide an effective alternative to parameter tuning for multimodal analysis.

\begin{table}[t]
  \centering
  \small
  \caption{Comparison of VQA paradigms on PaveBench, including zero-shot, LoRA fine-tuning, and the proposed agent-augmented framework. Results are reported on numerical and language metrics.}
  \label{tab:agent_metrics_comparison}
  \renewcommand{\arraystretch}{1.2}
  \resizebox{1.0\linewidth}{!}{
  \begin{tabular}{l ccc cccc}
  \toprule
  \multirow{2.8}{*}{VLM Model} & \multicolumn{3}{c}{Numerical Metrics (\%)} & \multicolumn{4}{c}{Language Metrics (\%)} \\
  \cmidrule(lr){2-4} \cmidrule(lr){5-8} 
   & \makecell{Cls.\\Acc. ($\uparrow$)} & \makecell{Loc.\\Token-F1 ($\uparrow$)} & \makecell{Quant.\\MAPE  ($\downarrow$)} & \makecell{ROUGE-L} ($\uparrow$) & BLEU ($\uparrow$) & METEOR ($\uparrow$) & \makecell{BERT\\Score ($\uparrow$)} \\ 
  \midrule
  \textbf{Qwen2.5-VL-3B}~\cite{wang2023qwenvl} & & & & & & & \\
  \quad Base (zero-shot) & 65.18 & 16.39 & 116.20 & 8.32 & 0.79 & 14.39 & 83.69 \\
  \quad + LoRA FT & \underline{88.24} & \textbf{43.77} & \underline{47.01} & \underline{52.40} & \underline{20.82} & \textbf{44.66} & \textbf{92.86} \\
  \quad + Agent Aug. (\textbf{ours}) & \textbf{89.68} & \underline{42.66} & \textbf{35.40} & \textbf{53.41} & \textbf{21.95} & \underline{44.22} & \underline{92.78} \\
  \midrule
  \textbf{DeepSeek-VL2-small}~\cite{lu2024deepseekvl} & & & & & & & \\
  \quad Base (zero-shot) & 55.48 & 40.36 & 531.21 & 24.82 & 4.77 & 23.89 & 88.17 \\
  \quad + LoRA FT & \textbf{92.98} & \textbf{59.23} & \underline{48.69} & \textbf{52.67} & \textbf{20.57} & \textbf{43.40} & \textbf{93.21} \\
  \quad + Agent Aug. (\textbf{ours}) & \underline{92.80} & \underline{42.08} & \textbf{26.91} & \underline{50.39} & \underline{18.08} & \underline{37.79} & \underline{92.65} \\
  \midrule
  \textbf{LLaVA-OneVision-7B}~\cite{li2024llavaonevision} & & & & & & & \\
  \quad Base (zero-shot) & 60.91 & 27.71 & 158.24 & 11.01 & 1.69 & 15.95 & 85.38 \\
  \quad + LoRA FT & \underline{83.04} & \textbf{55.19} & \underline{66.95} & \textbf{50.82} & \textbf{17.20} & \textbf{38.52} & \textbf{92.76} \\
  \quad + Agent Aug. (\textbf{ours}) & \textbf{83.33} & \underline{46.64} & \textbf{26.14} & \underline{46.09} & \underline{13.27} & \underline{33.35} & \underline{91.24} \\
  \bottomrule
  \end{tabular}
  }
\end{table}

\section{Conlusion}

In this paper, we present PaveBench, a unified benchmark for pavement distress analysis that extends from visual perception to multimodal reasoning. PaveBench provides high-resolution top-down imagery with unified annotations for classification, detection, and segmentation, and includes hard distractor scenarios to evaluate robustness under realistic conditions. Building on this visual foundation, we construct PaveVQA, a large-scale dataset with real images that supports multi-turn dialogue, quantitative queries, and expert-verified reasoning. To address the spatial and numerical limitations of general vision-language models, we further introduce an agent-augmented VQA framework. This framework routes user queries to specialized visual tools and separates semantic reasoning from geometric measurement, producing verifiable and visually grounded outputs. Extensive experiments demonstrate the effectiveness of both the dataset and the framework. PaveBench provides a foundation for moving pavement analysis from visual recognition toward more reliable and interactive understanding.

\bibliographystyle{ACM-Reference-Format}
\bibliography{references}

@String{Computer = "{IEEE} Computer" }

@String{Springer = "Springer-Verlag" }

@article{amhaz2015automatic,
  title={Automatic crack detection on two-dimensional pavement images: An algorithm based on minimal path selection},
  author={Amhaz, Rabih and Chambon, Sylvie and Idier, J{\'e}r{\^o}me and Baltazart, Vincent},
  journal={IEEE Transactions on Intelligent Transportation Systems},
  volume={17},
  number={10},
  pages={2718--2729},
  year={2016}
}

@article{maeda2019road,
  title={Road damage detection and classification using deep neural networks with smartphone images},
  author={Maeda, Hiroya and Sekimoto, Yoshihide and Seto, Toshikazu and Kashiyama, Takehiro and Omata, Hiroshi},
  journal={Computer-Aided Civil and Infrastructure Engineering},
  volume={33},
  number={12},
  pages={1127--1141},
  year={2018}
}

@inproceedings{maeda2021rdd2020,
  title={Generative adversarial network for road damage detection},
  author={Maeda, Hiroya and Kashiyama, Takehiro and Sekimoto, Yoshihide and Seto, Toshikazu and Omata, Hiroshi},
  booktitle={Computer-Aided Civil and Infrastructure Engineering},
  volume={36},
  number={1},
  pages={47--60},
  year={2021}
}

@inproceedings{liu2023visual,
  title={Visual Instruction Tuning},
  author={Liu, Haotian and Li, Chunyuan and Wu, Qingyang and Lee, Yong Jae},
  booktitle={Proceedings of Advances in Neural Information Processing Systems},
  pages={34892--34916},
  year={2023}
}

@inproceedings{masry2022chartqa,
  title={ChartQA: A Benchmark for Question Answering about Charts with Visual and Logical Reasoning},
  author={Masry, Ahmed and Long, Do Xuan and Tan, Jia Qing and Joty, Shafiq and Hoque, Enamul},
  booktitle={Proceedings of Association for Computational Linguistics},
  pages={2263--2279},
  year={2022}
}

@inproceedings{liu2021slake,
  title={SLAKE: A Semantically-Labeled Knowledge-Enhanced Dataset for Medical Visual Question Answering},
  author={Liu, Bo and Zhan, Li-Ming and Xu, Li and Ma, Lin and Yang, Yan and Wu, Xiaodan Mo},
  booktitle={Proceedings of IEEE International Symposium on Biomedical Imaging},
  pages={1650--1654},
  year={2021}
}

@article{zhang2023pmcvqa,
  title={PMC-VQA: Visual Instruction Tuning for Medical Visual Question Answering},
  author={Zhang, Xiaoman and Wu, Chaoyi and Zhao, Ziheng and others},
  journal={arXiv},
  pages={1--19},
  year={2023}
}

@inproceedings{li2023llavamed,
  title={LLaVA-Med: Training a Large Language-and-Vision Assistant for Biomedicine in One Day},
  author={Li, Chunyuan and Wong, Cliff and Zhang, Sheng and others},
  booktitle={Proceedings of Advances in Neural Information Processing Systems},
  pages={28541--28564},
  year={2023}
}

@inproceedings{yin2025toolvqa,
  title={ToolVQA: A Dataset for Multi-step Reasoning VQA with External Tools},
  author={Yin, Shaofeng and Lei, Ting and Liu, Yang},
  booktitle={Proceedings of IEEE/CVF International Conference on Computer Vision},
  pages={4424--4433},
  year={2025}
}

@inproceedings{xiao2026roadbench,
  title={Roadbench: A vision-language foundation model and benchmark for road damage understanding},
  author={Xiao, Xi and Zhang, Yunbei and Wang, Janet and Zhao, Lin and Wei, Yuxiang and Li, Hengjia and Li, Yanshu and Wang, Xiao and Roy, Swalpa Kumar and Xu, Hao and others},
  booktitle={Proceedings of the IEEE/CVF Winter Conference on Applications of Computer Vision},
  pages={6016--6026},
  year={2026}
}

@inproceedings{gupta2023visual,
  title={Visual programming: Compositional visual reasoning without training},
  author={Gupta, Tanmay and Kembhavi, Aniruddha},
  booktitle={Proceedings of IEEE/CVF International Conference on Computer Vision and Pattern Recognition},
  pages={14953--14962},
  year={2023}
}

@article{shi2016crackforest,
  author    = {Yong Shi and Limeng Cui and Zhiquan Qi and Fan Meng and Zhensong Chen},
  title     = {Automatic Road Crack Detection Using Random Structured Forests},
  journal   = {IEEE Transactions on Intelligent Transportation Systems},
  volume    = {17},
  number    = {12},
  pages     = {3434--3445},
  year      = {2016}
}

@article{yang2019feature,
  title={Feature pyramid and hierarchical boosting network for pavement crack detection},
  author={Yang, Fan and Zhang, Lei and Yu, Sijia and Prokhorov, Danil and Mei, Xue and Ling, Haibin},
  journal={IEEE transactions on intelligent transportation systems},
  volume={21},
  number={4},
  pages={1525--1535},
  year={2019},
}

@article{zou2012cracktree,
  author    = {Qingguo Zou and Yu Cao and Qingquan Li and Qingzhou Mao and Song Wang},
  title     = {CrackTree: Automatic Crack Detection from Pavement Images},
  journal   = {Pattern Recognition Letters},
  volume    = {33},
  number    = {3},
  pages     = {227--238},
  year      = {2012}
}

@misc{kaggle11k,
  title        = {Crack Segmentation Dataset},
  author       = {{Lakshay Middha}},
  year         = {2020},
  howpublished = {Kaggle dataset},
}

@inproceedings{eisenbach2017get,
  title={How to get pavement distress detection ready for deep learning? A systematic approach},
  author={Eisenbach, Markus and Stricker, Ronny and Seichter, Daniel and Amende, Karl and Debes, Klaus and Sesselmann, Maximilian and Ebersbach, Dirk and Stoeckert, Ulrike and Gross, Horst-Michael},
  booktitle={2017 international joint conference on neural networks (IJCNN)},
  pages={2039--2047},
  year={2017},
}

@article{arya2024rdd2022,
  title={RDD2022: A multi-national image dataset for automatic road damage detection},
  author={Arya, Deeksha and Maeda, Hiroya and Ghosh, Sanjay Kumar and Toshniwal, Durga and Sekimoto, Yoshihide},
  journal={Geoscience Data Journal},
  volume={11},
  number={4},
  pages={846--862},
  year={2024},
}

@article{liu2024pavedistress,
  title={PaveDistress: A comprehensive dataset of pavement distresses detection},
  author={Liu, Zhen and Wu, Wenxiu and Gu, Xingyu and Cui, Bingyan},
  journal={Data in Brief},
  volume={57},
  pages={111111},
  year={2024},
}

@article{liu2019deepcrack,
  title={DeepCrack: A deep hierarchical feature learning architecture for crack segmentation},
  author={Liu, Yahui and Yao, Jian and Lu, Xiaohu and Xie, Renping and Li, Li},
  journal={Neurocomputing},
  volume={338},
  pages={139--153},
  year={2019},
}

@article{majidifard2020pavement,
  title={Pavement image datasets: A new benchmark dataset to classify and densify pavement distresses},
  author={Majidifard, Hamed and Jin, Peng and Adu-Gyamfi, Yaw and Buttlar, William G},
  journal={Transportation Research Record},
  volume={2674},
  number={2},
  pages={328--339},
  year={2020},

}

@inproceedings{deng2009imagenet,
  title={Imagenet: A large-scale hierarchical image database},
  author={Deng, Jia and Dong, Wei and Socher, Richard and Li, Li-Jia and Li, Kai and Fei-Fei, Li},
  booktitle={Proceedings of IEEE/CVF International Conference on Computer Vision and Pattern Recognition},
  pages={248--255},
  year={2009},
  organization={Ieee}
}

@inproceedings{lin2014microsoft,
  title={Microsoft coco: Common objects in context},
  author={Lin, Tsung-Yi and Maire, Michael and Belongie, Serge and Hays, James and Perona, Pietro and Ramanan, Deva and Doll{\'a}r, Piotr and Zitnick, C Lawrence},
  booktitle={Proceedings of European Conference on Computer Vision},
  pages={740--755},
  year={2014},
  organization={Springer}
}

@inproceedings{cordts2016cityscapes,
  title={The cityscapes dataset for semantic urban scene understanding},
  author={Cordts, Marius and Omran, Mohamed and Ramos, Sebastian and Rehfeld, Timo and Enzweiler, Markus and Benenson, Rodrigo and Franke, Uwe and Roth, Stefan and Schiele, Bernt},
  booktitle={Proceedings of the IEEE conference on computer vision and pattern recognition},
  pages={3213--3223},
  year={2016}
}

@inproceedings{lin2004rouge,
  title={Rouge: A package for automatic evaluation of summaries},
  author={Lin, Chin-Yew},
  booktitle={Proceedings of Text Summarization Branches Out},
  pages={74--81},
  year={2004}
}

@inproceedings{papineni2002bleu,
  title={Bleu: a method for automatic evaluation of machine translation},
  author={Papineni, Kishore and Roukos, Salim and Ward, Todd and Zhu, Wei-Jing},
  booktitle={Proceedings of the 40th annual meeting of the Association for Computational Linguistics},
  pages={311--318},
  year={2002}
}

@inproceedings{banerjee2005meteor,
  title={METEOR: An automatic metric for MT evaluation with improved correlation with human judgments},
  author={Banerjee, Satanjeev and Lavie, Alon},
  booktitle={Proceedings of the acl workshop on intrinsic and extrinsic evaluation measures for machine translation and/or summarization},
  pages={65--72},
  year={2005}
}

@inproceedings{zhang2019bertscore,
  title={{BERTScore}: Evaluating Text Generation with {BERT}},
  author={Tianyi Zhang and Varsha Kishore and Felix Wu and Kilian Q. Weinberger and Yoav Artzi},
  booktitle={Proceedings of International Conference on Learning Representations},
  year={2020}
}

@inproceedings{woo2023convnext,
  title={Convnext v2: Co-designing and scaling convnets with masked autoencoders},
  author={Woo, Sanghyun and Debnath, Shoubhik and Hu, Ronghang and Chen, Xinlei and Liu, Zhuang and Kweon, In So and Xie, Saining},
  booktitle={Proceedings of IEEE/CVF International Conference on Computer Vision and Pattern Recognition},
  pages={16133--16142},
  year={2023}
}

@article{hatamizadeh2023fastervit,
  title={Fastervit: Fast vision transformers with hierarchical attention},
  author={Hatamizadeh, Ali and Heinrich, Greg and Yin, Hongxu and Tao, Andrew and Alvarez, Jose M and Kautz, Jan and Molchanov, Pavlo},
  journal={arXiv},
  pages={1--14},
  year={2023}
}

@inproceedings{zeng2025efficient,
  title={An Efficient Hybrid Vision Transformer for TinyML Applications},
  author={Zeng, Fanhong and Li, Huanan and Guan, Juntao and Fan, Rui and Wu, Tong and Wang, Xilong and Lai, Rui},
  booktitle={Proceedings of IEEE/CVF International Conference on Computer Vision},
  pages={19914--19924},
  year={2025}
}

@inproceedings{wang2025lsnet,
  title={Lsnet: See large, focus small},
  author={Wang, Ao and Chen, Hui and Lin, Zijia and Han, Jungong and Ding, Guiguang},
  booktitle={Proceedings of IEEE/CVF International Conference on Computer Vision and Pattern Recognition},
  pages={9718--9729},
  year={2025}
}

@inproceedings{lou2025overlock,
  title={Overlock: An overview-first-look-closely-next convnet with context-mixing dynamic kernels},
  author={Lou, Meng and Yu, Yizhou},
  booktitle={Proceedings of IEEE/CVF International Conference on Computer Vision and Pattern Recognition},
  pages={128--138},
  year={2025}
}

@inproceedings{huang2025deim,
  title={Deim: Detr with improved matching for fast convergence},
  author={Huang, Shihua and Lu, Zhichao and Cun, Xiaodong and Yu, Yongjun and Zhou, Xiao and Shen, Xi},
  booktitle={Proceedings of the computer vision and pattern recognition conference},
  pages={15162--15171},
  year={2025},
}

@inproceedings{nan2025mi,
  title={MI-DETR: an object detection model with multi-time inquiries mechanism},
  author={Nan, Zhixiong and Li, Xianghong and Dai, Jifeng and Xiang, Tao},
  booktitle={Proceedings of IEEE/CVF International Conference on Computer Vision and Pattern Recognition},
  pages={4703--4712},
  year={2025}
}

@inproceedings{li2025remdet,
  title={Remdet: Rethinking efficient model design for uav object detection},
  author={Li, Chen and Zhao, Rui and Wang, Zeyu and Xu, Huiying and Zhu, Xinzhong},
  booktitle={Proceedings of the AAAI conference on artificial intelligence},
  volume={39},
  number={5},
  pages={4643--4651},
  year={2025}
}

@inproceedings{chen2018encoder,
  title={Encoder-decoder with atrous separable convolution for semantic image segmentation},
  author={Chen, Liang-Chieh and Zhu, Yukun and Papandreou, George and Schroff, Florian and Adam, Hartwig},
  booktitle={Proceedings of European Conference on Computer Vision},
  pages={801--818},
  year={2018}
}

@inproceedings{xie2021segformer,
  title={{SegFormer}: Simple and efficient design for semantic segmentation with transformers},
  author={Xie, Enze and Wang, Wenhai and Yu, Zhiding and Anandkumar, Anima and Alvarez, Jose M and Luo, Ping},
  booktitle={Proceedings of Advances in Neural Information Processing Systems},
  volume={34},
  pages={12077--12090},
  year={2021}
}

@article{liu2025sosnet,
  title={{SOSNet}: Real-Time Small Object Segmentation via Hierarchical Decoding and Example Mining},
  author={Liu, Wang and Kang, Xudong and Duan, Puhong and Xie, Zhuojun and Wei, Xiaohui and Li, Shutao},
  journal={IEEE Transactions on Neural Networks and Learning Systems},
  volume={36},
  number={2},
  pages={3071--3083},
  year={2025}
}

@inproceedings{liu2025scsegamba,
  author       = {Hui Liu and
                  Chen Jia and
                  Fan Shi and
                  Xu Cheng and
                  Shengyong Chen},
  title        = {SCSegamba: Lightweight Structure-Aware Vision Mamba for Crack Segmentation
                  in Structures},
  booktitle    = {Proceedings of IEEE/CVF International Conference on Computer Vision and Pattern Recognition},
  pages        = {29406--29416},
  year         = {2025}
}

@article{wang2023qwenvl,
  title={Qwen-vl: A versatile vision-language model for understanding, localization},
  author={Bai, Jinze and Bai, Shuai and Yang, Shusheng and Wang, Shijie and Tan, Sinan and Wang, Peng and Lin, Junyang and Zhou, Chang and Zhou, Jingren},
  journal={Text Reading, and Beyond},
  volume={2},
  number={1},
  pages={1},
  year={2023}
}

@article{sapkota2025yolo26,
  title={YOLO26: key architectural enhancements and performance benchmarking for real-time object detection},
  author={Sapkota, Ranjan and Cheppally, Rahul Harsha and Sharda, Ajay and Karkee, Manoj},
  journal={arXiv},
  pages={1--15},
  year={2025}
}

@article{lu2024deepseekvl,
  title={DeepSeek-VL: towards real-world vision-language understanding},
  author={Lu, Haoyu and Liu, Wen and Zhang, Bo and Wang, Bingxuan and Dong, Kai and Liu, Bo and Sun, Jingxiang and Ren, Tongzheng and Li, Zhuoshu and Sun, Yaofeng and others},
  journal={arXiv},
  pages={1--29},
  year={2024}
}

@article{li2024llavaonevision,
  title={Llava-onevision: Easy visual task transfer},
  author={Li, Bo and Zhang, Yuanhan and Guo, Dong and Zhang, Renrui and Li, Feng and Zhang, Hao and Zhang, Kaichen and Li, Yanwei and Liu, Ziwei and Li, Chunyuan},
  journal={arXiv},
  pages={1--33},
  year={2024}
}

@inproceedings{hu2021lora,
  title={{LoRA}: Low-Rank Adaptation of Large Language Models},
  author={Edward J Hu and Yelong Shen and Phillip Wallis and Zeyuan Allen-Zhu and Yuanzhi Li and Shean Wang and Lu Wang and Weizhu Chen},
  booktitle={Proceedings of International Conference on Learning Representations},
  pages={1--12},
  year={2022}
}

@inproceedings{yao2022react,
  title={React: Synergizing reasoning and acting in language models},
  author={Yao, Shunyu and Zhao, Jeffrey and Yu, Dian and Du, Nan and Shafran, Izhak and Narasimhan, Karthik R and Cao, Yuan},
  booktitle={The eleventh international conference on learning representations},
  year={2022}
}

@article{wu2023visualchatgpt,
  title={Visual {ChatGPT}: Talking, Drawing and Editing with Visual Foundation Models},
  author={Wu, Chenfei and Yin, Shengming and Qi, Weizhen and Wang, Xiaodong and Tang, Zecheng and Duan, Nan},
  journal={arXiv},
  pages={1--14},
  year={2023}
}
\end{document}